\documentstyle{article}

\title{PARSING A FLEXIBLE WORD ORDER LANGUAGE}

\author{Vladimir Pericliev and Alexander Grigorov\\
Institute of Mathematics,\\
Acd. G. Bonchev Str., bl.8, 1113 Sofia, Bulgaria,\\
E-mail: peri@bgearn.bitnet and grigorov@bgearn.bitnet }

\date{}

\begin{document}

\maketitle

\begin{abstract}
A logic formalism is presented which increases the expressive power of the
ID/LP format of GPSG by enlarging the inventory of ordering relations and
extending the domain of their application to non-siblings.  This allows a
concise, modular and declarative statement of intricate word order
regularities.
\end{abstract}

\section{Introduction}

Natural languages exhibit significant word order (WO) variation and intricate
ordering rules.  Despite the fact that specific languages show less variation
and complexity in such rules (e.\ g.\ those characterized by either fixed, or
totally free, WO), the vast majority of world languages lie somewhere
in-between these two extremes (e.\ g.\ {\it Steele 1981}).  Importantly, even
the proclaimed examples of rigid WO languages (English) exhibit variation,
whereas those with proclaimed total scrambling (Warlpiri; cf. {\it Hale 1981})
show restrictions ({\it Kashket 1987}).  Therefore, we need general grammar
formalism, capable of processing "flexible" WO (i.e.  complex WO regularities,
including both extremes).

There seem to be a number of requirements that such a formalism should (try to)
fulfil (e.\ g.\ {\it Pericliev and Grigorov 1992}).  Among these stand out the
formalism's:
\begin{itemize}

\item[(i)]
{\it Expressive power}, i.\ e.\ capability of (reasonably) handling complex WO
phenomena, or "flexible" WO.

\item[(ii)]
{\it Linguistic felicity}, i.\ e.\ capability of stating concisely and
declaratively WO rules in a way maximally approximating linguistic parlance in
similar situations.

\item[(iii)]
{\it Modularity}, i.\ e.\ the separation of constituency rules from the rules
pertaining to the linearization of these constituents (for there may be many,
and diverse, reasons for wanting linearization (and constituency) rules easily
modifiable, incl.\ the transparency of WO statements, the imprecision of our
current knowledge of ordering rules or the wish to tailor a system to a domain
with specific WO).

\item[(iv)]
{\it Reversibility}, i.\ e.\ the ability of a system to be used for both
parsing and generation (the reason being that, even if the system is originally
intended for a parser, complex WO rules may be conveniently tested in the
generation mode; in this sense it is not incidental that e.\ g.\ {\it Kay \&
Karttunen 1984} have first constructed a generator, and used it as a tool in
testing the (WO) rules of their grammar, and only then have converted it into a
parser).
\end{itemize}

In this paper, we present a logic-based formalism which attempts to satisfy the
above requirements.  A review shows that most previous approaches to WO within
the logic grammars paradigm ({\it Dahl \& Abramson 1990}) have not been
satisfactory.  Definite Clause Grammar, DCG, ({\it Pereira \& Warren 1980}),
with their CF-style rules, are not modular (in the sense above), so will have
to specify explicitly each ordering of constituents in a separate rule, which
results in an intolerably great number of rules in parsing a free WO language
(e.\ g.\ for 5 constituents, which may freely permute, the number of rules is
$5!  = 120$).  Other approaches center around the notion of a "gap" (or
"skip").  In Gapping Grammar (GG), for instance ({\it Dahl \& Abramson 1984},
esp.\ Dahl 1984), where a rule with a gap may be viewed as a meta-rule,
standing for a set of CF rules, free WO is more economically expressed,
however, due the unnaturalness of expressing permutations by gaps, GGs
generally are clumsy for expressing flexible WO, WO is not declaratively and
modularly expressed, and GGs cannot be used for generation (being besides not
efficiently implementable).  Another powerful formalism, Contextual
Discontinuous Grammar ({\it Saint-Dizier 1988}), which overcomes the GGs
problems with generative capacity, is also far from being transparent and
declarative in expressing WO (e.\ g.\ rules with fixed WO are transformed into
free order ones by introducing special rules, containing symbols with no
linguistic motivation, etc.).

\section{Problems for the ID/LP format}

In the Immediate Dominance/Linear Precedence (ID/LP) format of GPSG ({\it
Gazdar \& Pullum 1981}, {\it Gazdar et al.\ 1985}), where the information,
concerning constituency ($=$ immediate dominance) and linear order, is
separated, WO rules are concisely, declaratively and modularly expressed over
the domain of local-trees (i.\ e.\ trees of depth 1).  E.\ g.\ the ID rule
{\bf A $\rightarrow_{ID}$ B, C, D}, if no linearization restrictions are
declared,stands for the mother node expanded into its siblings appearing in
any order;declaring the restriction \{ {\bf D $<$ C} \} e.\ g., it stands for
the CFG rules \{ {\bf A $\rightarrow$ B D C}, {\bf A $\rightarrow$ D B C} and
{\bf A $\rightarrow$ D C B} \}.

It is important to note that in GPSG the linear precedence rules stated for a
pair of sibling constituents should be valid for the whole set of grammar rules
in which these constituents occur, and not just for some specific rule (this
"global" empirical constraint on WO is called the Exhaustive Constant Partial
Ordering (ECPO) property).

However, there are problems with ECPO.  They may be illustrated with a simple
example from Bulgarian.  Consider a grammar describing sentences with a
reflexive verb and a reflexive particle (the NP-subject and the adverb being
optional), responsible for expressions whose English equivalent is e.\ g.\
"(Ivan) shaved himself (yesterday)".
\begin{itemize}

\item[(1)]
{\bf S $\rightarrow_{ID}$ NP, VP}

\item[(2)]
{\bf S $\rightarrow_{ID}$ VP} \hspace{2.0cm} \% omitted subject

\item[(3)]
{\bf VP $\rightarrow_{ID}$ V[refl], Part[refl], Adv}

\item[(4)]
{\bf VP $\rightarrow_{ID}$ V[refl], Part[refl]} \hspace{2.0cm} \% omitted
adverb

\end{itemize}

First, assume we derive a sentence, applying rules (2) and (3).  (5a-b) are the
only acceptable linearizations of the sister constituents in (3).
\begin{itemize}

\item[(5a)]
\begin{tabular}{ccc}
Brasna ({\bf V[refl]})  &  se ({\bf Part[refl]})  &  vcera ({\bf Adv}) \\
shaved                  &  himself                &  yesterday
\end{tabular}

\item[(5b)]
\begin{tabular}{ccc}
Vcera ({\bf Adv})  &  se ({\bf Part[refl]})  &  brasna ({\bf V[refl]}) \\
Yesterday          &  himself                &  shaved
\end{tabular}\\
(meaning: (Someone) shaved himself yesterday)
\end{itemize}
LP rules however cannot enforce exactly these orderings because the CFG,
corresponding to (5a-b), viz.
\begin{itemize}
\item[(6)]
{\bf A $\rightarrow$ B C D} \\
{\bf A $\rightarrow$ D C B}
\end{itemize}
is non-ECPO.  Thus, fixing any ordering between any two constituents in (3)
will, of necessity, block at least one of the correct orderings (5a-b);
alternatively, sanctioning no WO restriction will result in overgeneration,
admitting, besides the grammatical (5a-b), 4 ungrammatical permutations.  This
inability to impose an arbitrary ordering on siblings we will call the
ordering-problem of ID/LP grammars.

Now assume we derive a sentence, applying rules (1) and (4).  The ordering of
the siblings, reflexive verb and particle, in (4) now depends on the order of
nodes {\bf NP} and {\bf VP} higher up in the tree in rule (1):  if {\bf NP}
precedes {\bf VP} in (1), then the reflexive particle must precede the verb in
(4), otherwise it should follow it.
\begin{itemize}

\item[(7a)]
\begin{tabular}{ccc}
Ivan ({\bf NP})  &  se ({\bf Part[refl]})  &  brasna ({\bf V[refl]}) \\
Ivan             &  himself                &  shaved
\end{tabular}

\item[(7b)]
\begin{tabular}{ccc}
Brasna ({\bf V[refl]})  &  se ({\bf Part[refl]})  &  Ivan ({\bf NP}) \\
Shaved                  &  himself                &  Ivan
\end{tabular}\\
(meaning: Ivan shaved himself)
\end{itemize}
Again we are in trouble since LP rules cannot impose orderings among
non-siblings, their domain of application being just siblings.  This we call
the {\it domain-problem} of ID/LP grammars.  It is essential to note that the
domain-problem may not be remedied (even if we are inclined to sacrifice
linguistic intuitions) by "flattening" the tree, e.\ g.\ collapsing rules (1)
and (4) into
\begin{itemize}
\item[(8)]
{\bf S $\rightarrow_{ID}$ NP, V[refl], Part[refl]}
\end{itemize}
Escaping the second problem, thrusts us into the first:  we now cannot properly
order the siblings, the CFG, corresponding to (7a-b), being the non-ECPO (6).

Sporadic counter-evidence for ECPO grammars has been found for some languages
like English (the verb-particle construction, {\it Sag 1987}, {\it Pollard and
Sag 1987}), German (complex fronting, {\it Uszkoreit 1985}, {\it Engelkamp et
al.  1992}) and Finnish (the adverb myos 'also, too' {\it Zwicky and Nevis
1986}).  Bulgarian offers massive counter-evidence ({\it Pericliev 1992b}); one
major example, the Bulgarian clitic system, we discuss in Section 4.

\section{The formalism}

EFOG (Extended Flexible word Order Grammar) extends the expressive power of the
ID/LP format.  First, EFOG introduces further WO restrictions in addition to
precedence (enabling it to avoid the ordering-problem), and, second, the
formalism extends the domain of application of these WO restrictions (in order
to handle the domain-problem).

In the immediate dominance part of rules EFOG has two types of constituents:
{\it non-contiguous} (notated:  {\bf \#Node}) and contiguous (notated just:
{\bf Node}), where {\bf Node} is some node.  Informally, a contiguous node
shows that its daughters form a contiguous sequence, whereas a non-contiguous
one allows its daughters to be interspersed among the sisters of this
non-contiguous node.  E.\ g.\ in EFOG notation (using a double arrow for ID
rules, small case letters for constants and upper case ones for variables), the
grammar of the Latin sentence: {\it Puella bona puerum parvum amat (good girl
loves small boy)}, grammatical in all its 120 permutations and, besides, having
discontinuity in the noun phrases, we capture with the following structured
EFOG rules with no WO restrictions:
\begin{quote}
\bf
s $\Rightarrow$ \#np(nom), \#vp. \\
np(Case) $\Rightarrow$ adj(Case), noun(Case). \\
vp $\Rightarrow$ verb, \#np(acc).
\rm
\end{quote}
accompanied by the dictionary rules:
\begin{quote}
\bf
verb $\Rightarrow$ [amat]. \\
adj(nom) $\Rightarrow$ [bona]. \\
adj(acc) $\Rightarrow$ [parvum]. \\
noun(nom) $\Rightarrow$ [puella]. \\
noun(acc) $\Rightarrow$ [puerum].
\end{quote}
The non-contiguous nodes allow us to impose an ordering (or to intersperse, as
in the above case) all their daughter nodes without having to sacrifice the
natural constituencies.  It will be clear that this extension of the domain of
LP rules (which can go any depth we like), besides ordering between
non-siblings, allows an elegant treatment of discontinuities.

In order to solve the ordering-problem, we have introduced additional WO
constraints.  The following atomic WO constraints have been defined:
\begin{itemize}
\item
{\it Precedence constraints}:
\begin{itemize}
\item
precedes (e.\ g.\ {\bf a} $<$ {\bf b})
\item
immediately precedes ({\bf a} $<<$ {\bf b}) (we also maintain the notation, $>$
and $>>$, for (immediately) follows; see commentary below)
\end{itemize}
\item
{\it Adjacency constraints}:
\begin{itemize}
\item
is adjacent ({\bf a} $<>$ {\bf b}).
\end{itemize}
\item
{\it Position constraints}:
\begin{itemize}
\item
is positioned first/last (e.\ g.\ {\bf first(a, Node)}, where {\bf Node} is a
node; e.\ g.\ {\bf first(a, s)} designates that {\bf a} is sentence-initial.
\end{itemize}
\end{itemize}

We also allow atomic WO constraints to combine into complex logical
expressions, using the following operators with obvious semantics:
\begin{itemize}
\item
Conjunction (notated: {\bf and})
\item
Disjunction ({\bf or})
\item
Negation ({\bf not})
\item
Implication ({\bf if}, e.\ g.\ {\bf (b $>>$ a) if (a $<$ c)} )
\item
Equivalence ({\bf iff}, e.\ g.\ {\bf (b $>>$ a) iff (a $<$ c)} )
\item
Ifthenelse ({\bf ifthenelse})
\end{itemize}

Our WO restriction language is, of course, partly logically redundant (e.\ g.\
immediately precedence may be expressed through precedence and adjacency, and
so is the case with the last two of the operators, etc.).  However, what is
logically is not necessarily psychologically equivalent, and our goal has been
to maintain a linguist-friendly notation (cf.\ requirement (ii) of Section 1).
To take just one example, we have 'after' in addition to 'before', since
linguists normally speak of precedence of dependent with respect to head word,
not vice versa, and hence will use both expressions in respective situations
(surely it is not by chance that NLs also have both words).

As a simple example of the ordering possibilities of EFOG, consider the WO
Universal 20 (of Greenberg and Hawkins) to the effect that NPs comprising
dem(onstrative), num(eral), adj(ective) and noun can appear in that order, or
in its mirror-image.  We can write a "universal" rule enforcing adjacent
permutations of all constituents as follows:
\begin{quote}
\bf
np $\Rightarrow$ dem, num, adj, noun. \\
lp: dem $<>$ num and num $<>$ adj and adj $<>$ noun.
\rm
\end{quote}

\section{Bulgarian clitics}

Bulgarian clitics fall into different categories:
\begin{itemize}
\item[(1)]
nominals (short accusative pronouns:  {\it me} "me", {\it te} "you", etc.;
short dative pronouns:  {\it mi} "to me", {\it ti} "to you", etc.);
\item[(2)]
verbs (the present tense forms of "to be" {\it sam} "am", {\it si} "(you) are",
etc.);
\item[(3)]
adjectives (short possessive pronouns:  {\it mi} "my", {\it ti} "your", etc.;
short reflexive pronoun:  {\it si} "one's own");
\item[(4)]
particles (interrogative {\it li} "do", reflexive {\it se}
"myself/yourself\ldots", the negative {\it ne} "no(t)", etc.).
\end{itemize}
They have the distribution of the specific categories they belong to, but show
diverse, and quite complex orderings, varying in accordance with the positions
of their siblings/non-siblings as well as the position of other clitics
appearing in the sentence.
\footnote{This often results in discontinuities (or non-projectivities).  For
an automated way of discovering and a description of such constructs in
Bulgarian, cf.\ {\it Pericliev and Ilarionov 1986}, and {\it Pericliev 1986}.}
In effect, their ordering as a rule cannot be correctly stated in the standard
ID/LP format.

By way of illustration, below we present the EFOG version (simplified for
expository reasons) of the grammar (1-4) from Section 2 to get the flavour of
how we handle the problems mentioned there.  The ID rules are as follows (note
that the non-contiguous node {\bf \#vp} allows its daughters {\bf v(refl)},
{\bf part(refl)}, and {\bf adv} to be ordered with respect to {\bf np}):
\begin{itemize}
\item[(1')]
{\bf s $\Rightarrow$ np, \#vp.}
\item[(2')]
{\bf s $\Rightarrow$ vp.} \hspace{2.0cm} \% omitted subject
\item[(3')]
{\bf vp $\Rightarrow$ v(refl), part(refl), adv.}
\item[(4')]
{\bf vp $\Rightarrow$ v(refl), part(refl).} \hspace{2.0cm}
\% omitted adverb \\[3pt]
{\bf np $\Rightarrow$ [ivan].}\\
{\bf v(refl) $\Rightarrow$ [brasna].}\\
{\bf part(refl) $\Rightarrow$ [se].}\\
{\bf adv $\Rightarrow$ [vcera].}
\end{itemize}

The WO of {\bf v(refl)} and {\bf part(refl)} is as follows.  First, the
reflexive particle never occurs sentence-initially (information we cannot
express in ID/LP); in EFOG we express this as:
\begin{quote}
{\bf lp: not(first(part(refl),s)).}
\end{quote}
Secondly, we use the default rule 'ifthenelse' to declare the regularity that
the particle in question immediately precedes the verb, unless when the verb
occurs sentence-initially, in which case the particle immediately follows it
(which is of course also inexpressible in ID/LP):
\begin{quote}
{\bf lp: ifthenelse(first(v(refl),s),}\\
\hspace*{3.0cm}{\bf v(refl) $<<$ part(refl),}\\
\hspace*{3.0cm}{\bf part(refl) $<<$ v(refl)).}
\end{quote}

These two straightforward LP rules thus are all we need to get exactly the
linearizations we want:  those of (5a-b) and (7a-b), as well as all and the
only other correct expressions derivable from the ID grammar.  These LP rules
are also interesting in that they express the overall behaviour of a number of
other proclitically behaving clitics (as e.g.  those with nominal and verbal
nature; see above).

Because of space limitations we cannot enter into further details here.
Suffice it to say that EFOG was tested successfully in the description of this
very complicated domain \footnote{ For the difficulties in handling the
adjectival clitics in pure DCG, cf.\ {\it Pericliev 1992a}.}
as well as in some other hard ordering problems in Bulgarian.

\section{Conclusion}

Logic grammars have generally failed to handle flexible WO in a satisfactory
way.  We have described a formalism which allows the grammar-writer to express
complex WO rules in a language (including discontinuity) in a concise, modular
and natural way.  EFOG extends the expressive power of the ID/LP format in both
allowing complex LP rules and extending their domain of application.

EFOG is based on a previous version of the formalism, called FOG ({\it
Pericliev and Grigorov 1992}), also seeking to overcome the difficulties with
the ID/LP format.  FOG however looked for different solutions to the problems
(e.\ g.\ using LP rules attached to each specific ID rule, rather than global
ones, which unnecessarily proliferated the LP part of the grammar; or employing
flattening rather than having non-contiguous grammar symbols to the same
effect).  EFOG is also related to FO-TAG ({\it Becker et al.  1991}) and the
HPSG approach ({\it Engelkamp et al.\ 1992}, {\it Oliva 1992}) in extending the
domain of applicability of LP rules.  A comparisson with these formalisms is
beyond the scope of this study; we may only mention here that our inventory of
LP relations is larger, and unlike e.\ g.\ the latter approach we do not
confine to binary branching trees.

\section*{References}

\begin{description}
\item
Becker T., A.\ Joshi and O.\ Rambow (1991).\ Long-distance scrambling and TAG.\
{\it Fifth Conference of the EACL}, Berlin, pp.\  21-26.
\item
Dahl, V.\ (1984).\ More on Gapping Grammars.\ {\it Proc.\ of the Intern.\
Conf.\ on 5th Generation Computer Systems}, ICOT, pp.\ 669-677.
\item
Dahl, V.\ and H.\ Abramson (1984).\ On Gapping Grammars.\ {\it Proc.\ 2nd
Intern.\ Conf.\ on Logic Programming}, Uppsala, pp.\ 77-88.
\item
Dahl, V.\ and H.\ Abramson (1990).\ {\it Logic Grammars}.\ Springer.
\item
Engelkamp, J., G.\ Erbach and H.\ Uszkoreit (1992).\ Handling linear
precedence constraints by unification.\ {\it Annual Meeting of the ACL}.
\item
Gazdar, G.\ and G.\ Pullum (1981).\ Subcategorization, constituent order and
the notion of "head".\ M.\ Moortgat et al.\ (eds.) {\it The Scope of Lexical
Rules}, Dordrecht, Holland, pp.\ 107-123.
\item
Gazdar, G., E.\ Klein, G.\ Pullum and I.\ Sag (1985).\ {\it Generalized Phrase
Structure Grammar}.\ Harvard, Cambr., Mass.
\item
Hale, K.\ (1983).\ Warlpiri and the grammar of non-configurational languages.\
{\it Natural Language and Linguistic Theory}, 1, pp.\ 5-49.
\item
Kashket, M.\ (1987).\ A GB-based parser for Warlpiri, a free-word order
language.\ MIT AI Laboratory.
\item
Kay, M.\ and L.\ Karttunen (1984).\ Parsing a free word order language.\ D.\
Dowty et al.\ (eds.) {\it Natural Language Parsing}.\ The Cambridge ACL
series.
\item
Oliva, K.\ (1992).\ Word order constraints in binary branching syntactic
structures.\ University of Saarland Report (appearing also in {\it COLING'92}).
\item
Pereira, F.C.N.\ and D.H.D.\ Warren (1980).\ Definite Clause Grammars for
Natural Language Analysis.\ {\it Artificial Intelligence}, v.13, pp.\ 231-278.
\item
Pericliev, V.\ (1986).\ Non-projective con-structions in Bulgarian.\ {\it 2nd
World Congress of Bulgaristics}, Sofia, pp.\ 271-280 (in Bulgarian).
\item
Pericliev, V.\ and I.\ Ilarionov (1986).\ Testing the projectivity hypothesis.\
{\it COLING'86}, Bonn, pp.\ 56- 58.\
\item
Pericliev, V.\ (1992a).\ A referent grammar treatment of some problems in the
Bulgarian nominal phrase.\ {\it Studia Linguistica}, Stockholm, pp.\ 49-62.
\item
Pericliev, V.\ (1992b).\ The ID/LP format:  counter-evidence from Bulgarian,
(ms).
\item
Pericliev, V.\ and A.\ Grigorov (1992).\ Extending Definite Clause Grammar to
handle flexible word order.\ B.\ du Boulay et al.\ (eds.) {\it Artificial
Intelligence V}, North Holland, pp.\ 161-170.
\item
Pollard C., I.\ Sag (1987).\ {\it Information-Based Syntax and Semantics}.\
Vol.\ 1:  Fundamentals.\ CSLI Lecture Notes No.\ 13, Stanford, CA.
\item
Sag, I.\ (1987).\ Grammatical hierarchy and linear precedence.\ {\it Syntax and
Semantics}, v.20, pp.\ 303- 339.
\item
Saint-Dizier, P.\ (1988).\ Contextual Discon-tinuous Grammars.\ {\it Natural
Language Understanding and Logic Programming}, II, North Holland, pp.\ 29-43.
\item
Steele, S.\ (1981).\ Word order variation:  a typological study.\ G.\ Greenberg
(ed.) {\it Universals of Language}, v.4, Stanford.
\item
Uszkoreit, H.\ (1985).\ Linear precedence in discontinuous constituents:
complex fronting in German.\ SRI International, Technical Note 371.
\item
Zwicky, A.\ (1986).\ Immediate precedence in GPSG.\ {\it OSU WPL 32}, pp.\
133-138.
\end{description}

\end{document}